\documentclass[
]{ceurart}

\usepackage[frozencache=true,cachedir=minted-cache]{minted}
\setminted{breaklines=true}

\usepackage{algorithm}
\usepackage{algorithmicx}
\usepackage{algpseudocode}

\begin{document}

\copyrightyear{2024}
\copyrightclause{Copyright for this paper by its authors.
  Use permitted under Creative Commons License Attribution 4.0
  International (CC BY 4.0).}

\conference{IAL@ECML-PKDD'24:
  8\textsuperscript{th} Intl. Worksh. \& Tutorial on Interactive Adaptive Learning,
  Sep. 9\textsuperscript{th}, 2024, Vilnius, Lithuania}

\title{Deep Transfer Hashing for Adaptive Learning on Federated Streaming Data}
\subtitle{Sample. Hash. Adapt. Repeat.}


\author[1,2,3]{Manuel Röder}[%
orcid=0009-0003-4907-3999,
email=manuel.roeder@thws.de
]
\cormark[1]

\address[1]{Faculty of Computer Science and Business Information Systems, TUAS
Würzburg-Schweinfurt, Würzburg, Germany}
\address[2]{Faculty of Technology, Bielefeld University, Bielefeld, Germany}
\address[3]{Center for Artificial Intelligence and Robotics Würzburg, Germany}

\author[1]{Frank-Michael Schleif}[%
orcid=0000-0002-7539-1283,
email=frank-michael.schleif@thws.de
]

\cortext[1]{Corresponding author.}

\begin{abstract}
This extended abstract explores the integration of federated learning with deep transfer hashing for distributed prediction tasks, emphasizing resource-efficient client training from evolving data streams.
Federated learning allows multiple clients to collaboratively train a shared model while maintaining data privacy \-- by incorporating deep transfer hashing, high-dimensional data can be converted into compact hash codes, reducing data transmission size and network loads.
The proposed framework utilizes transfer learning, pre-training deep neural networks on a central server, and fine-tuning on clients to enhance model accuracy and adaptability.
A selective hash code sharing mechanism using a privacy-preserving global memory bank further supports client fine-tuning.
This approach addresses challenges in previous research by improving computational efficiency and scalability.
Practical applications include Car2X event predictions, where a shared model is collectively trained to recognize traffic patterns, aiding in tasks such as traffic density assessment and accident detection.
The research aims to develop a robust framework that combines federated learning, deep transfer hashing and transfer learning for efficient and secure downstream task execution.
\end{abstract}


\maketitle

\section{Introduction and Background}

The rapid growth of data and the increasing emphasis on privacy-preserving machine learning techniques have spurred significant interest in \emph{federated learning} (FL)~\cite{pmlr-v54-mcmahan17a}.
This extended abstract explores the integration of FL with  \emph{deep transfer hashing} (DTH)~\cite{Zhou2018deeptransfer} methods for distributed downstream classification and retrieval tasks, focusing on resource-aware FL client training from \emph{evolving data streams}~\cite{marfoq2023federated} and leveraging transfer learning through pre-training deep neural network models on the FL server and employ the learned model weights for client model initialization.
In addition, the client fine-tuning process is further supported by a selective hash code sharing mechanism through the use of a globally available but privacy preserving memory bank.
The overarching goal of this concept paper is to present and elaborate on a novel idea that addresses challenges identified in previous research~\cite{roder2023efficient, roder2024crossing} and to initiate further discussions. 

FL is a distributed machine learning paradigm where multiple clients collaboratively train a shared model while keeping their data decentralized and secure.
This approach is particularly beneficial for applications that require strict data protection and security measures.
By combining federated learning with deep transfer hashing techniques, we aim to efficiently convert high-dimensional data into compact, low-dimensional hash codes, significantly decreasing data transmission size between the FL server and clients, reducing network transfer loads, and potentially improving client inference efficiency.
Deep transfer hashing methods have proven to be highly effective in reducing the dimensionality of data while preserving its intrinsic structure.
This capability is crucial for classification tasks, where the high-dimensional nature of data often poses significant computational challenges.
In this context, \emph{locality sensitive hashing} and \emph{learning to hash} approaches have been widely used.
However, traditional locality sensitive hashing methods require the construction and administration of numerous hash tables, which can be impractical for distributed optimisation tasks such as those observed in FL.
Learning to hash potentially provides a more scalable and efficient solution by leveraging the powerful representation capabilities of deep learning models to learn complex hash functions in an end-to-end manner.
To further enhance the performance of our proposed framework, we employ transfer learning by pre-training the DNN model on the high-performance server.
This pre-trained model can then be fine-tuned on client devices using their local data streams, ensuring that the model adapts to the specific characteristics of each client's data.
This approach not only accelerates the training process but also improves the overall accuracy and generalization capability of the model.

\begin{figure}
  \label{fig_overview}
  \centering
  \includegraphics[width=\linewidth]{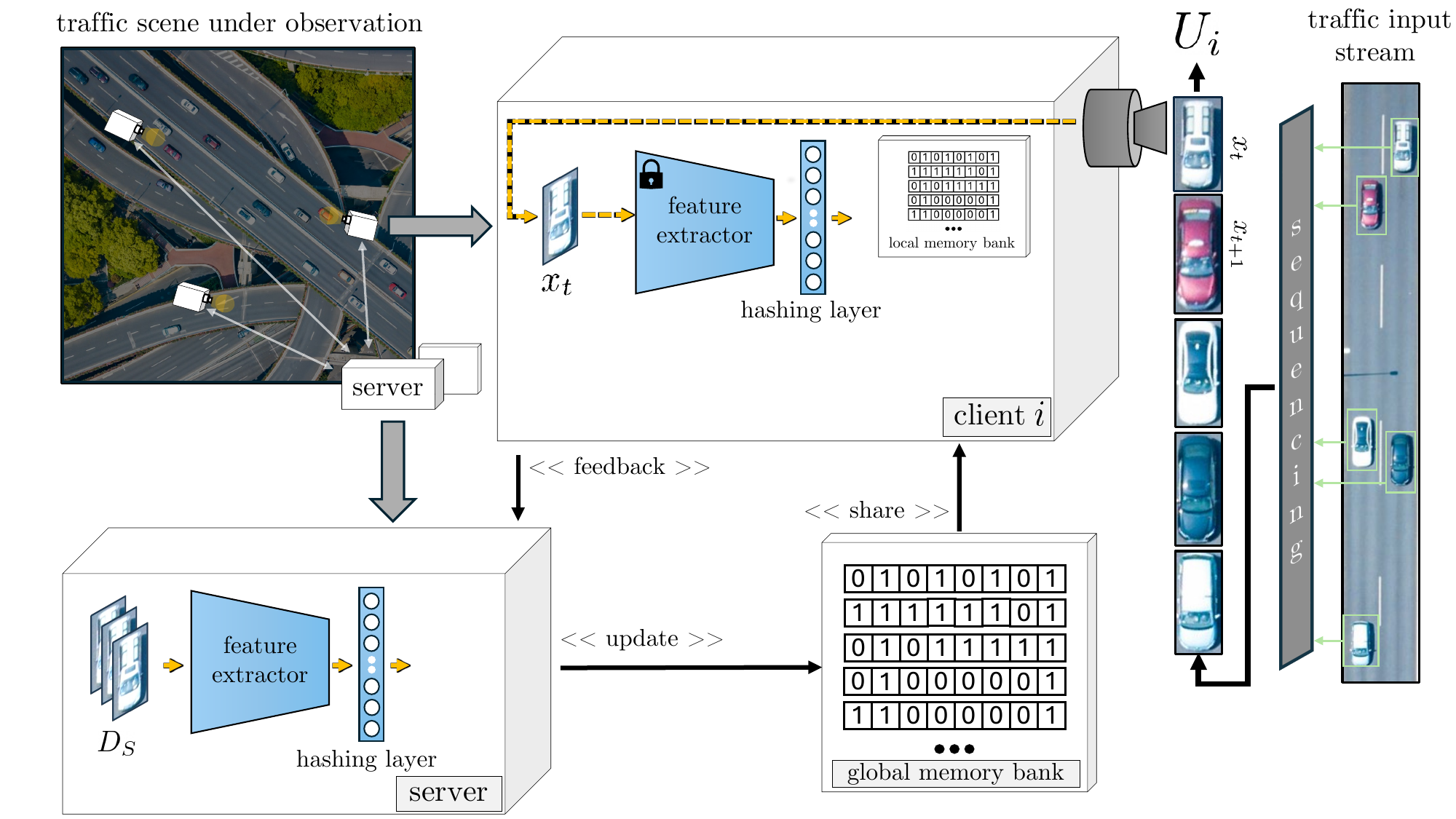}
  \caption{Car2x scenario: FL clients with monitoring sensors (camera, radar) sample from a traffic input stream at different geographic locations. The central server instance maintains a global memory bank that is accessible for all client devices to support local client model fine-tuning. The feature extractor and hashing layer are pre-trained on the large-scale server data set $D_S$. The central server administrates a global memory bank.}
\end{figure}

We aim to integrate our approach in practical scenarios that involve raw or pre-processed data points from monitoring sensors installed at various locations, such as traffic cameras and surveillance cameras as seen in Figure~\ref{fig_overview}.
In these scenarios, models can learn to recognize patterns, objects, or anomalies.
For example, in a \emph{Car2X}~\cite{car2x2024} driven application, our concept can support various use-case areas such as Intersection Movement Assist, Intersection Collision Avoidance or Green Light Optimal Speed Advisory by enabling models to distinguish between vehicle types, assess real-time traffic density and detect and alert about accidents.
In summary, our research aims to combine the strengths of FL, DTH and transfer learning to develop a robust and efficient framework for downstream classification and retrieval tasks, while adhering to the constraints imposed by FL. 

\section{Methodology}
We consider a FL environment composed of a central, resource-heavy server $S$ tasked with network orchestration and multiple resource-restricted clients indexed by $i$, where $i=1, \ldots I$ as outlined in Fig.~\ref{fig_overview}.
Additionally, client $i$ inspects sample $x_t$, seen only once, from the \emph{non-I.I.D.} data stream $U_i$ at time $t$, in which the occurrence of objects is not evenly distributed, using an arbitrary sensing device.

In the preparation phase, a task-specific deep neural network model is pre-trained on the server to learn a task-specific hash function $h$ that maps an input \( x \) to a binary code \( b \), facilitating the nearest neighbor search used for prediction inference. 
A well-designed hash function should preserve the relative distances between items in the original space, meaning that items close to a specific query in Hamming space should also be close to the query in the original space~\cite{xiao_survey2023}.
Subsequently, the FL server distributes the learned model weights at the start of a new FL round and also initializes the global memory bank with learned hash codes, raising \hyperlink{oq:one}{Open Question 1}.
The establishment of a global memory bank, which is fed with hash codes by the FL server in a sophisticated manner, offers enhanced data protection on the one hand and a reduced network data flow on the other, as only data that is intended to support local model training is made available.
Each FL client participating in the distributed learning process follows a simple distributed \textbf{SHAR} pattern  for local model adaptation as outlined in Algorithm~\ref{alg:shar}:

\noindent \textit{\textbf{S}ample. }
Recall that the FL client $i$ samples data point $x_t$ from the data stream $U_i$ at time $t$. The selection of the sampling algorithm depends on various parameters like the downstream task, the quality of the streamed data, the underlying model and the cost of sample labeling.
The authors of this work already proposed an \emph{Active Learning}-based sampling method ``that identifies relevant stream observations to optimize the underlying client model, given a local labeling budget, and performs instantaneous labeling decisions without relying on
any memory buffering strategies''~\cite{roedersparse2024}. Hence, the framework of this paper is not bound or limited to intelligent sampling strategies and also works with heuristic approaches like for example the selection of samples based on fixed time intervals.

\noindent \textit{\textbf{H}ash.}
Participating clients receive the pre-trained model and have on-demand access to the global memory bank $M_S$, utilizing these resources to perform localized fine-tuning with the sampled data set.
This process involves each client further optimizing the hash function $h_i$ to better adapt to their specific data, thereby enhancing the model’s performance for their particular fine-tuning tasks.
To achieve this, we aim to employ a pointwise-based hashing method.
Recent advancements typically construct the classification loss within the Hamming space. Specifically, these methods generate a set of central hash codes, each associated with a specific class label.
The objective is to enforce the network outputs to converge towards their respective hash centers using various loss terms, thereby ensuring that the hash function preserves the relative distances between items effectively~\cite{xiao_survey2023} Our intention is to support the local hash code learning step by enriching the adaptation phase with relevant information obtained from the global memory bank $M_S$, raising \hyperlink{oq:two}{Open Question 2}.

\noindent \textit{\textbf{A}dapt.}
An important consideration in deploying a FL model for critical real-world applications like Car2X is the phenomenon of \emph{concept drift}, where the statistical properties of data points sampled from the client data stream change over time.
This drift can result from evolving user preferences, seasonal variations, or other dynamic factors influencing the data distribution.
To maintain the efficacy of our transferred hashing algorithm, it is crucial to evaluate and implement strategies for detecting concept drift.
Integrating mechanisms to handle the changing data problem into our incremental hash code learning process ensures that the model adapts to new patterns and continues to generate accurate and relevant hash codes.
Effective detection and adaptation techniques will help to maintain the performance and reliability of the model, even as the underlying data distributions change over time, raising \hyperlink{oq:three}{Open Question 3}.

\noindent \textit{\textbf{R}epeat. } The adaptation on local clients is repeated until the global model converges.
Updates received on the server (hash codes, model parameters) from the clients after each round of federated training must be integrated into both the server model and the global memory bank and thereby maintaining data integrity and data privacy, raising \hyperlink{oq:four}{Open Question 4}.

Overall, evaluation and benchmarking of the proposed method is essential, with a clear justification for its preference over existing techniques.
A major challenge is the availability of ground truth data, especially to assess the model's handling of concept drift, raising \hyperlink{oq:five}{Open Question 5}.

\begin{algorithm}
    \caption{\small \textbf{S}ample. \textbf{H}ash. \textbf{A}dapt. \textbf{R}epeat.\label{alg:shar}}
    \begin{algorithmic}[1]
        \Require Pre-trained hash function $h_i$ on client $i$, unlabeled stream of samples $U_i$, global memory bank $M_S$
        \State Initialize global memory bank $M_S$
        \For{\textbf{each} federated learning round}
            \State Initialize $t = 1$
            \State Initialize $B_i = \{\}$ \Comment{Initialize local hash code storage}
            \For{$x_t \in U_i$} \Comment{\textbf{Sample} from data stream}
                \State $b_t = \texttt{\textsc{calculate\_hash}}(x_t)$ \Comment{\textbf{Hash} code generation}
                \State \hspace*{\algorithmicindent} $B_i \leftarrow B_i \cup \{b_t\}$
                \State $t \leftarrow t + 1$
            \EndFor
            \State $h_i  =$ \texttt{\textsc{adapt}}($B_i$, $M_S$) \Comment{\textbf{Adapt} client model}
            \State Send update to server $S$
            
        \EndFor \Comment{\textbf{Repeat} until convergence}
    \State \Return Converged global learning model, educated global memory bank $M_S$
    \end{algorithmic}
\end{algorithm}

\section{Conclusion and Open Questions}
In this work, we presented the integration of federated learning with deep transfer hashing methods for distributed classification and retrieval tasks.
Our approach focuses on resource-aware client training from evolving data streams, leveraging transfer learning through pre-trained models on the FL server, and utilizing selective hash code sharing via a privacy-preserving global memory bank.
This integration is supposed to efficiently convert high-dimensional data into compact hash codes, reducing network data transmission and improving client training and inference efficiency.

This concept paper outlines a foundational idea aimed at addressing challenges in data privacy, computational efficiency, and scalability in FL prediction tasks.
To underline the importance and relevance of our research, we identified and outlined a real-world use-case that enhances several areas of Car2X application.
We seek to initiate further discussions and collaborations to refine these concepts and advance privacy-preserving machine learning techniques.
By attending this workshop, we hope to gain educated insights from the discussions on the following open questions:

\noindent
\hypertarget{oq:one}{\textbf{Open Question 1: }} What hash codes to include in the global memory bank (class prototypes) and what is the best type and structure of the memory layout (map, tree, graph)?\\
\hypertarget{oq:two}{\textbf{Open Question 2: }}  How can external memory banks improve distributed FL learning while sticking to all FL constraints (data privacy, communication efficiency, …) ?\\
\hypertarget{oq:three}{\textbf{Open Question 3: }}  How to properly integrate the incremental aspect in deep transfer hashing to take account for concept drift as proposed in ~\cite{tian2023} within a non-FL environment? \\
\hypertarget{oq:four}{\textbf{Open Question 4: }}  How to adapt and integrate external memory management strategies (e.g. from continual learning, reservoir sampling)?\\
\hypertarget{oq:five}{\textbf{Open Question 5: }}  How can we evaluate the proposed approach to demonstrate its unique advantages, and what specific conditions must be met for its successful application?\\

\begin{acknowledgments}
  MR is supported through the Bavarian HighTech Agenda, specifically by the Würzburg Center for Artificial Intelligence and Robotics (CAIRO) and the ProPere THWS scholarship.  
\end{acknowledgments}

\bibliography{ECML2024_WS}

\begin{thebibliography}{9}
\expandafter\ifx\csname natexlab\endcsname\relax\def\natexlab#1{#1}\fi
\providecommand{\url}[1]{\texttt{#1}}
\providecommand{\href}[2]{#2}
\providecommand{\path}[1]{#1}
\providecommand{\DOIprefix}{doi:}
\providecommand{\ArXivprefix}{arXiv:}
\providecommand{\URLprefix}{URL: }
\providecommand{\Pubmedprefix}{pmid:}
\providecommand{\doi}[1]{\href{http://dx.doi.org/#1}{\path{#1}}}
\providecommand{\Pubmed}[1]{\href{pmid:#1}{\path{#1}}}
\providecommand{\bibinfo}[2]{#2}
\ifx\xfnm\relax \def\xfnm[#1]{\unskip,\space#1}\fi
\bibitem[{McMahan et~al.(2017)McMahan, Moore, Ramage, Hampson, and Arcas}]{pmlr-v54-mcmahan17a}
\bibinfo{author}{B.~McMahan}, \bibinfo{author}{E.~Moore}, \bibinfo{author}{D.~Ramage}, \bibinfo{author}{S.~Hampson}, \bibinfo{author}{B.~A.~y. Arcas},
\newblock \bibinfo{title}{Communication-efficient learning of deep networks from decentralized data},
\newblock in: \bibinfo{editor}{A.~Singh}, \bibinfo{editor}{J.~Zhu} (Eds.), \bibinfo{booktitle}{Proceedings of the 20th international conference on artificial intelligence and statistics}, volume~\bibinfo{volume}{54} of \textit{\bibinfo{series}{Proceedings of machine learning research}}, \bibinfo{year}{2017}, pp. \bibinfo{pages}{1273--1282}. \URLprefix \url{https://proceedings.mlr.press/v54/mcmahan17a.html}.
\bibitem[{Zhou et~al.(2018)Zhou, Zhao, Peng, Fang, Qin, and Goh}]{Zhou2018deeptransfer}
\bibinfo{author}{J.~T. Zhou}, \bibinfo{author}{H.~Zhao}, \bibinfo{author}{X.~Peng}, \bibinfo{author}{M.~Fang}, \bibinfo{author}{Z.~Qin}, \bibinfo{author}{R.~S.~M. Goh},
\newblock \bibinfo{title}{Transfer hashing: From shallow to deep},
\newblock \bibinfo{journal}{IEEE Transactions on Neural Networks and Learning Systems} \bibinfo{volume}{29} (\bibinfo{year}{2018}) \bibinfo{pages}{6191--6201}. \DOIprefix\doi{10.1109/TNNLS.2018.2827036}.
\bibitem[{Marfoq et~al.(2023)Marfoq, Neglia, Kameni, and Vidal}]{marfoq2023federated}
\bibinfo{author}{O.~Marfoq}, \bibinfo{author}{G.~Neglia}, \bibinfo{author}{L.~Kameni}, \bibinfo{author}{R.~Vidal}, \bibinfo{title}{Federated learning for data streams}, \bibinfo{year}{2023}. \href{http://arxiv.org/abs/2301.01542}{{\tt arXiv:2301.01542}}.
\bibitem[{Röder et~al.(2023)Röder, Heller, Münch, and Schleif}]{roder2023efficient}
\bibinfo{author}{M.~Röder}, \bibinfo{author}{L.~Heller}, \bibinfo{author}{M.~Münch}, \bibinfo{author}{F.-M. Schleif}, \bibinfo{title}{Efficient cross-domain federated learning by mixstyle approximation}, \bibinfo{year}{2023}. \href{http://arxiv.org/abs/2312.07064}{{\tt arXiv:2312.07064}}.
\bibitem[{Röder et~al.(2024)Röder, Münch, Raab, and Schleif}]{roder2024crossing}
\bibinfo{author}{M.~Röder}, \bibinfo{author}{M.~Münch}, \bibinfo{author}{C.~Raab}, \bibinfo{author}{F.-M. Schleif},
\newblock \bibinfo{title}{Crossing domain borders with federated few-shot adaptation},
\newblock in: \bibinfo{booktitle}{Proceedings of the 13th International Conference on Pattern Recognition Applications and Methods - Volume 1: ICPRAM}, \bibinfo{organization}{INSTICC}, \bibinfo{publisher}{SciTePress}, \bibinfo{year}{2024}, pp. \bibinfo{pages}{511--521}. \DOIprefix\doi{10.5220/0012351900003654}.
\bibitem[{C2C-CC(2024)}]{car2x2024}
\bibinfo{author}{C2C-CC}, \bibinfo{title}{{Car 2 Car Communication Concortium}}, \bibinfo{year}{2024}. \URLprefix \url{https://www.car-2-car.org/}.
\bibitem[{Luo et~al.(2023)Luo, Wang, Wu, Chen, Deng, Huang, and Hua}]{xiao_survey2023}
\bibinfo{author}{X.~Luo}, \bibinfo{author}{H.~Wang}, \bibinfo{author}{D.~Wu}, \bibinfo{author}{C.~Chen}, \bibinfo{author}{M.~Deng}, \bibinfo{author}{J.~Huang}, \bibinfo{author}{X.-S. Hua},
\newblock \bibinfo{title}{A survey on deep hashing methods},
\newblock \bibinfo{journal}{ACM Trans. Knowl. Discov. Data} \bibinfo{volume}{17} (\bibinfo{year}{2023}). \URLprefix \url{https://doi.org/10.1145/3532624}. \DOIprefix\doi{10.1145/3532624}.
\bibitem[{Röder and Schleif(2024)}]{roedersparse2024}
\bibinfo{author}{M.~Röder}, \bibinfo{author}{F.-M. Schleif}, \bibinfo{title}{{Sparse Uncertainty-Informed Sampling from Federated Streaming Data}}, \bibinfo{year}{2024}. \bibinfo{note}{Accepted for publication in proceedings of \emph{European Symposium on Artificial Neural Networks, Computational Intelligence and Machine Learning} 2024}.
\bibitem[{Tian et~al.(2023)Tian, Ng, and Xu}]{tian2023}
\bibinfo{author}{X.~Tian}, \bibinfo{author}{W.~W.~Y. Ng}, \bibinfo{author}{H.~Xu},
\newblock \bibinfo{title}{Deep incremental hashing for semantic image retrieval with concept drift},
\newblock \bibinfo{journal}{IEEE Transactions on Big Data} \bibinfo{volume}{9} (\bibinfo{year}{2023}) \bibinfo{pages}{1102--1115}. \DOIprefix\doi{10.1109/TBDATA.2022.3233457}.

\end{thebibliography}

\end{document}